\documentclass[fleqn,10pt]{wlscirep}
\usepackage{helvet}
\usepackage[font=normal]{caption}
\usepackage[utf8]{inputenc}
\usepackage[T1]{fontenc}
\usepackage{comment}
\usepackage{amsmath}
\title{Automatized self-supervised learning for skin lesion screening}

\author[1,2]{Vullnet Useini}
\author[2,3]{Stephanie Tanadini-Lang}
\author[1]{Quentin Lohmeyer}
\author[1]{Mirko Meboldt}
\author[2,3]{Nicolaus Andratschke}
\author[4]{Ralph P. Braun}
\author[2,3]{Javier Barranco Garc\'ia}
\affil[1]{Department of Mechanical and Process Engineering, ETH Zurich, Leonhardstrasse 21,8092 Zurich, Switzerland.}
\affil[2]{Department of Radiation Oncology, University Hospital Zurich, R\"amistrasse 100, 8091, Zurich, Switzerland.}
\affil[3]{University of Zurich, R\"amistrasse 71, 8006, Zurich, Switzerland.}
\affil[4]{Department of Dermatology, University Hospital Zurich, Gloriastrasse 31, 8091, Zurich, Switzerland.}


\begin{abstract}
Melanoma, the deadliest form of skin cancer, has seen a steady increase in incidence rates worldwide, posing a significant challenge to dermatologists. Early detection is crucial for improving patient survival rates. However, performing total body screening (TBS), i.e., identifying suspicious lesions or ugly ducklings (UDs) by visual inspection, can be challenging and often requires sound expertise in pigmented lesions. To assist users of varying expertise levels, an artificial intelligence (AI) decision support tool was developed. Our solution identifies and characterizes UDs from real-world wide-field patient images. It employs a state-of-the-art object detection algorithm to locate and isolate all skin lesions present in a patient’s total body images. These lesions are then sorted based on their level of suspiciousness using a self-supervised AI approach, tailored to the specific context of the patient under examination. A clinical validation study was conducted to evaluate the tool’s performance. The results demonstrated an average sensitivity of 95\% for the top-10 AI-identified UDs on skin lesions selected by the majority of experts in pigmented skin lesions. The study also found that the tool increased dermatologists’ confidence when formulating a diagnosis, and the average majority agreement with the top-10 AI-identified UDs reached 100\% when assisted by our tool. With the development of this AI-based decision support tool, we aim to address the shortage of specialists, enable faster consultation times for patients, and demonstrate the impact and usability of AI-assisted screening. Future developments will include expanding the dataset to include histologically confirmed melanoma and validating the tool for additional body regions.
\end{abstract}
\begin{document}

\flushbottom
\maketitle

\thispagestyle{empty}

\section*{Introduction}\label{s:Introduction}
Malignant melanoma, the most serious form of skin cancer, is known for its ability to metastasize and rapidly spread to other organs. However, if diagnosed early enough, it can be removed through surgical intervention. Projections worldwide indicate an expected increase of 50\% in incidence and 68\% in the death rate by 2040~\cite{Arnold2022-hv}. This scenario, combined with a shortage of dermatologists to meet such demand, requires new solutions to assist experts in fighting this growing pandemic. Accurate diagnosis of skin cancer requires specific skills and experience that can only be acquired through proper training, limiting the number of professionals who can perform it effectively. A study funded by the Swiss Cancer League ~\cite{Badertscher2011} demonstrated that dermatological training significantly improved the performance of general practitioners (GPs) in diagnosing skin cancer. However, this improvement was temporary ~\cite{Badertscher2015-uc}, with its benefits fading within 12 months following the intervention. To assist dermatologists in the most time-consuming and error-prone task, total body screening (TBS), it is crucial to develop a reliable tool with high diagnostic accuracy. Such a tool would not only sustain long-term improvement but also enable the involvement of additional non-experts, such as nurses, technicians, or GPs. Artificial Intelligence (AI) has garnered significant interest in the analysis of dermoscopic images, as evidenced by competitions organized by the ISIC foundation ~\cite{ISIC} and numerous high-impact publications. A benchmark study in the field \cite{Esteva2017} underscores the effectiveness of AI-based tools in diagnosing skin cancer through dermoscopic imaging of individual lesions. Despite surpassing board-certified dermatologists in certain scenarios, the application of AI in routine consultations still faces substantial challenges, particularly when performing TBS and incorporating the patient’s context into the analysis. During TBS, the expert visually inspects the entire body and selects a few suspicious lesions or ugly ducklings (UDs) for further dermoscopic evaluation. The concept of a UD, a potential melanoma candidate that deviates from the individual’s nevus pattern, was first introduced in ~\cite{Grob1998}. However, it is only recently that AI-based research studies have begun to address the screening process by identifying and isolating suspicious lesions or UDs, including the patient’s context, as demonstrated in \cite{doi:10.1126/scitranslmed.abb3652}. This pioneering study presents a significant limitation regarding generalization and real-world application, as it relied on experienced dermatologists to manually label every single skin lesion to train the model. Given that each patient’s context is unique and manual labeling is both costly and time-consuming, it is clear that an alternative, labeling-independent approach is necessary. To address this issue, a proposal by \cite{VAEUDS}  suggests the use of a Variational Autoencoder (VAE) for outlier (ugly duckling) detection. This approach has several identified limitations. These include the non-uniformity of the dataset, which was obtained from multiple sources, and the pretraining of the Variational Autoencoder (VAE) on a small patient data pool, which could potentially introduce bias into the predictions. Additionally, validating the algorithm against a single dermatologist is suboptimal. These concerns highlight issues such as limited generalizability, a lack of sample diversity, the potential for single participant bias, and limited error detection.

To tackle these limitations, we propose a self-supervised architecture that is trained on real-world data (RWD). This data was acquired during routine consultations at the Dermatology Clinic of the University Hospital Zurich (USZ). Furthermore, we suggest validating this approach against multiple experienced dermatologists, who are currently considered the gold standard for melanoma diagnosis. Lastly, we present a clinical validation that involves diverse groups of interest. The main contributions of this study can be summarized as follows:

\begin{itemize}
    \item Use of real-world data of total body screening (RWD TBS) acquired in a standardized manner during routine consultations at the Dermatology Clinic of the USZ for training, validation and testing;
    \item Adoption and development of a self-supervised approach for UD sign characterization based on analysis of a single patient’s context;
    \item The clinical validation of the algorithm’s predictions against various groups of interest, and an evaluation of its impact as an AI-based decision support tool for melanoma screening;
\end{itemize}

\section*{Methods}
A self-supervised AI algorithm was developed for the automatic detection and characterization of UDs in high-risk patients (i.e., those with more than 100 nevi lesions). Established collaboration with the Dermatology Clinic of the USZ and the UZH Clinical Research Priority Program, provided access to a real-world dataset and medical experts. These resources were essential for the reliable clinical validation of AI-assisted total-body screening.
For the sake of simplicity, we have broken down the complete pipeline into several distinct stages. First, wide-field images were acquired during routine consultations at the USZ Dermatology Clinic. Second, all individual skin lesions were identified and extracted. Finally, the extracted lesions were transformed into a meaningful representation in which similar lesions were grouped together and scored based on their similarity to the average appearance of the lesions. In the final step, the AI pipeline was validated against medical experts.

\subsection*{End-to-end pipeline description}
\begin{figure}[ht]
\centering
\includegraphics[width=1\linewidth]{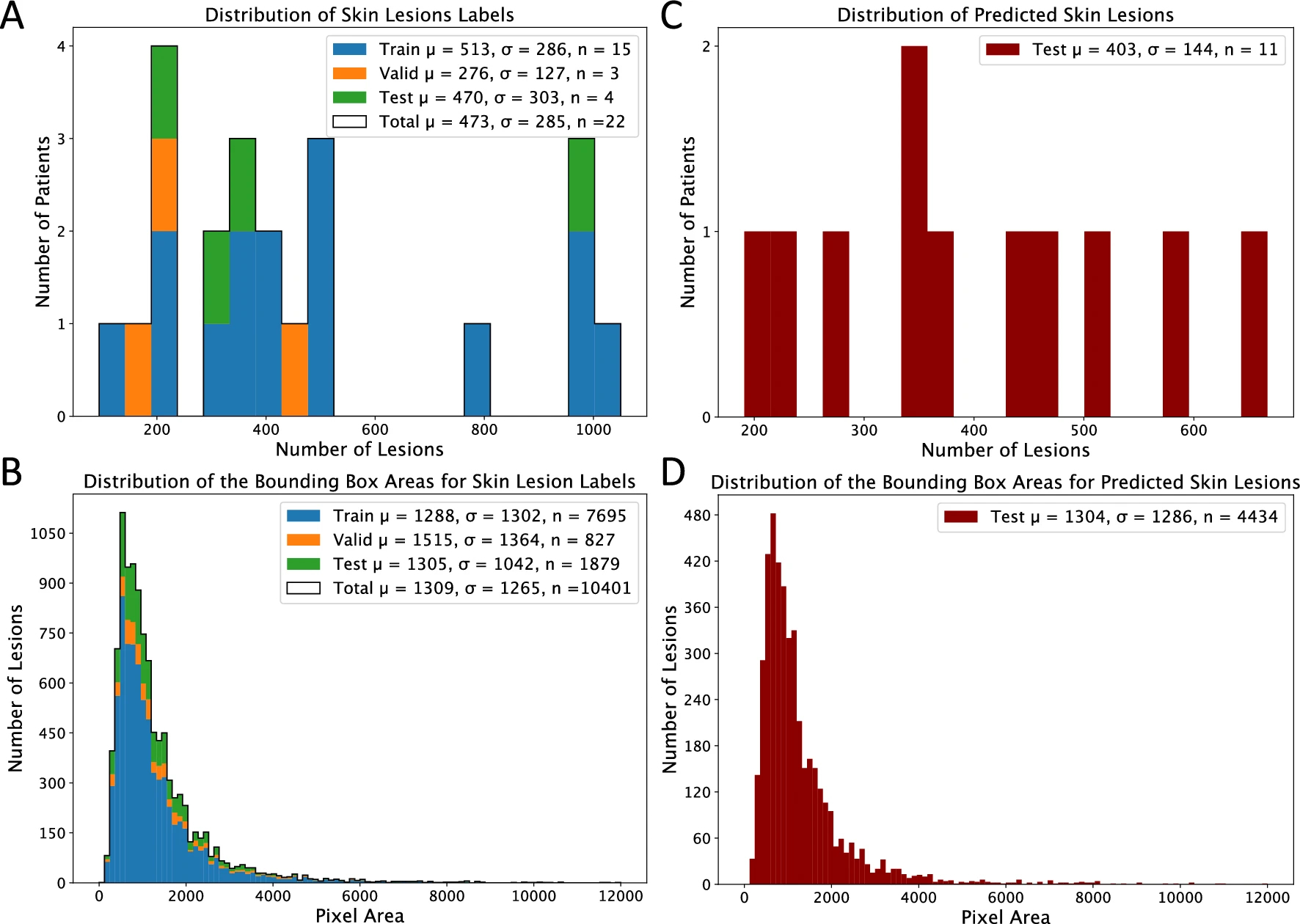}
\caption{(\textbf{A}) Patient-wise distribution of skin lesions. (\textbf{B}) Lesion-wise pixel area distribution of the labels created for the skin lesion detector. (\textbf{C}) Patient-wise distribution of skin lesions. (\textbf{D}) Lesion-wise pixel area distribution based on post-processed predictions for 11 unseen test patients.}\label{fig:Figure1}
\end{figure}

\textbf{Data acquisition, analysis and preprocessing}
\\
The recruitment process was conducted at the USZ Dermatology Clinic. All participants provided informed consent, specifically for this research study, which was approved by the Swiss Ethics Committee under project number 2020-01937. We used FotoFinder\copyright\ ATBM devices to semi-automatically capture high-resolution, polarized, total-body images. For this study, we acquired a total of 90 full-body dossiers from high-risk patients at the Dermatology Clinic of the USZ. To maintain patient confidentiality, these images were anonymized to contain only the dorsal region of each patient. Additionally, patients’ names were coded to further ensure privacy.
Due to time constraints for clinicians, we preprocessed only a randomly chosen subset of 33 total body images from the original set of 90. Out of these 33 refined dorsal images, 11 were randomly selected for the evaluation of our end-to-end pipeline. The remaining 22 images served as the data source for training the supervised skin lesion detection algorithm. These 22 images were randomly divided into subsets: 15 images for training, 3 for validation, and 4 for testing. While the data cannot be released due to privacy considerations, we present some descriptions in \autoref{fig:Figure1}. \hyperref[fig:Figure1]{\autoref{fig:Figure1}A} and \hyperref[fig:Figure1]{C} show histograms representing the number of lesions per patient in our labels batch for the skin lesion detection and test batch for the clinical validation study, respectively. Although the labels batch was limited to 22 subjects, the distribution of the number of lesions is similar across splits, with only the validation set representing a lower mean due to the absence of subjects with more than 500 lesions (\hyperref[fig:Figure1]{\autoref{fig:Figure1}A}). \hyperref[fig:Figure1]{\autoref{fig:Figure1}B} and \hyperref[fig:Figure1]{D} depict the pixel areas of the respective bounding boxes for each batch, showing that all evaluation splits have similar, right-tailed distributions.

\subsubsection*{Skin lesion detector}
To identify UDs within a patient’s context, it is essential to first identify all lesions. This task can be modeled as an object detection problem. For our specific application, we initially favored one-stage detection models due to their real-time applicability. Our model selection strategy prioritized high inference speed, with a contingency plan to shift to two-stage detection models if accuracy was compromised. Among the available options, YOLOR\cite{yolor} emerged as the architecture that best fit our criteria, as outlined below and benchmarked against the general MS COCO dataset. Our criteria included reliance solely on existing training data, optimized inference for high-resolution imagery, superior accuracy in detecting small objects measured through Average Precision for Small Objects (APS), and a backbone architecture adept at identifying intricate local patterns inherent in Convolutional Neural Networks (CNNs).

To tailor the official YOLOR architecture to our specific needs, we implemented strategic modifications. These involved restructuring the model architecture from 80 target classes to a single class, ensuring precise alignment with our application’s context. We customized aspect ratios for anchor boxes using K-nearest neighbors (KNN) algorithms derived from our unique training data. Additionally, we disabled color channel normalization to prevent undesired predictions within shadowed areas, increased the maximum allowable bounding boxes per image for heightened sensitivity, and deactivated the Non-Maximum Suppression (NMS) mechanism during inference mode, a deliberate choice made in the pre-merging phase. Furthermore, in our efforts to tailor the YOLOR architecture to meet our unique requirements, we utilized the official GitHub repository provided by the author, incorporating the previously mentioned modifications.

In the first step, we split our dataset into training, validation, and test sets. The next step consisted of creating labels for the supervised model. After consulting with a medical expert, it was decided to label anything that could potentially be considered a skin lesion, such as freckles, since it is difficult even for experts to know in advance what might be of interest. Additionally, it is critical to provide the AI algorithm with as much patient context as possible, to be able to accurately characterize the UDs.

The labeling process can be quite lengthy in patients with many skin lesions. To speed up the labeling process, a blob detector such as the one from OpenCV\cite{opencv_library} was used before starting the manual process.

One of the challenges in our application is detecting small objects within high-resolution images. Most deep learning approaches scale down the images to lower resolution to make them computationally feasible. However, this method usually results in a loss of information due to the lower input resolution. The common practice to overcome this problem in aerial imagery\cite{aerialliterature}, also adopted by others\cite{PrimaryOBDliterature}, is to create overlapping regions of the original images and labels before sending them to the AI model for training or inference. This approach was also followed in our study to ensure that all skin lesions were detected.

After successful training and optimization of our object detection model, the predictions from the extracted regions are merged back together for testing, and NMS is applied to the aggregated results, with an Intersection over Union (IoU) threshold of 10\%. The final version of the model was trained for 900 epochs on the aforementioned subset of 15 patients, validated on 3, and tested on 4. We calculated test results using established object detection metrics, including Recall, Precision, Average Recall (AR) and Average Precision (AP). The training was done using a Tesla®P100 16GB GPU, provided by the Swiss National Supercomputing Center (CSCS), without any pre-trained weights initialization and required 9 h to complete. After successfully training our model, we applied it to the data batch of 11 unseen patients, specifically reserved for the clinical validation study. Together with the UD labels created by the participants during the clinical validation study, these predictions enabled us to evaluate further the model's capability to detect skin lesions using a score metric that is less biased towards the more frequently appearing freckles, placing greater emphasis on UDs.

Finally, we implemented a filtering process to eliminate unwanted lesions, particularly those that were poorly illuminated as they influenced the UD selection by generating high outlier values. We excluded these lesions for the sake of this study, however they could be captured from different perspectives under better illumination conditions in future practical applications. A probabilistic approach was used to identify and exclude poorly illuminated lesions. This was achieved by utilizing a hand-crafted feature, specifically the mean intensity level of the frame pixels of the lesions. Given that the predicted bounding boxes always encompassed a portion of the skin surrounding the lesions, the frame pixels served as an ideal candidate for detecting suboptimal illumination conditions. The histograms of this feature generally exhibited a left-tailed Gaussian distribution. The probabilistic method was set up to identify lesions below a two-sigma event from their mean as poorly illuminated. This led to an average discard rate of 1.8\% for lesions. These lesions were predominantly located in shadowed areas of patients, such as the armpits or shoulder edges. These findings were obtained using the 3 validation patients from the batch of 22 patients.

\subsubsection*{Ugly duckling detector}

The primary objective was to develop an informative feature representation for the skin lesions extracted from the 11 unseen patients. This process was designed to mimic the workflow of dermatologists, who rely on comparing lesions based on their similarity. To effectively sort these identified skin lesions by their level of suspicion, it was necessary to embed them into a latent space that could provide a meaningful representation reflective of lesion similarity. In this context, distinguishing UDs required embedding them further away from the majority of lesions within this latent space. Utilizing a self-supervised approach proved instrumental in addressing the challenges outlined in our introduction. This approach was particularly effective in mitigating biases arising from mislabeled data or the context of the training patients, given the absence of publicly available labeled total body images. We explored various potential models to achieve this objective, emphasizing two main criteria: limited model parameters due to training solely on lesions from our test patient to circumvent contextual biases, and high accuracy in K-nearest neighbors (KNN) image classification tasks on broad datasets like ImageNet. Considering these requirements, DINO was selected for our representation task with ResNet18 as the backbone architecture \cite{DINO}.

In deep learning approaches, the input dimensions must be identical for each data instance. If this is not the case, the model scales the dimensions to match. Most of our extracted lesions have a square shape. However, in the other cases, we need to ensure that there is no scaling taking place that could mislead the UDs selection, for example, by changing the width to height ratio. As a solution, constant padding using the mean frame pixel values of the shorter dimension was proposed for lesions that do not have a width to height ratio of 1.

The next phase of the self-supervised approach for UDs detection involved the initiation of a pretext task to generate pseudo labels. To align with our task requirements, several modifications to the data augmentation processes outlined in the official DINO publication were made. First, we ensured that each lesion image was upscaled to 224x224 pixels, thereby preventing any lesions from being downscaled. Next, we employed random brightness jittering to counteract further poor illumination conditions. This was followed by the adoption of DINO’s general augmentation strategy, which includes random rescaling and cropping. Finally, to increase the significance of color differences, we adjusted the rescaling of each color channel by a factor of 10.

\begin{figure}[ht]
\centering
  \includegraphics[width=1\linewidth]{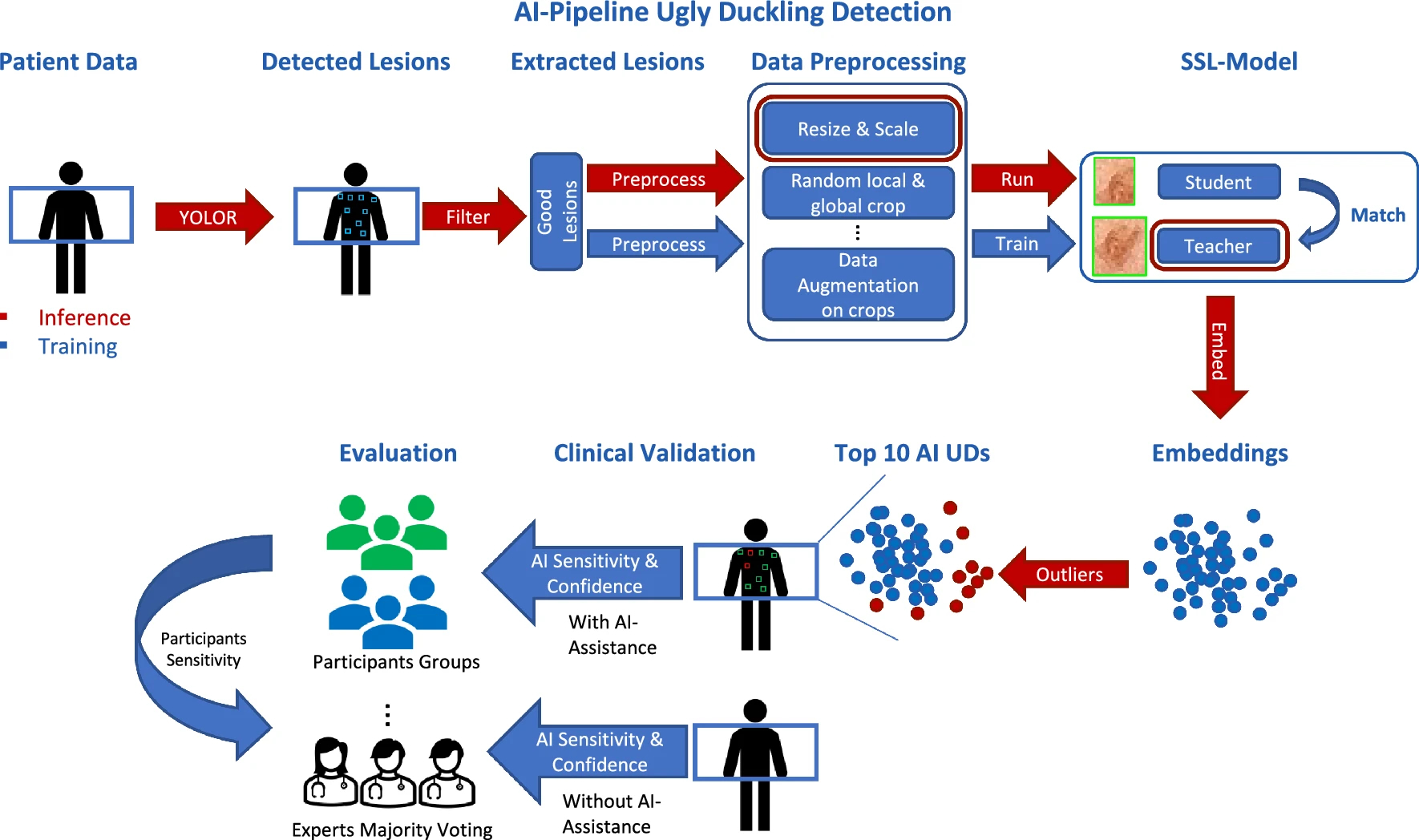}
  \caption{Overview of the proposed AI-pipeline for self-supervised ugly duckling selection in wide-field imagery (top) and clinical evaluation (bottom).}\label{fig:Figure2}
\end{figure}

The model was explicitly trained on the test patient’s lesions for a minimum of 200 epochs to ensure a robust initial convergence. This approach was adopted to avoid potential biases that could emerge from the contexts of different patients. For instance, some patients may predominantly have darker lesions, while others may have lighter or unusually shaped ones. These variations can influence the model’s learning and prediction capabilities to generalize for an unseen case. By training the model exclusively on the test patient’s lesions, we eliminate this bias, ensuring that the model’s predictions are tailored specifically to their unique lesion characteristics. This comprehensive approach enhances the model’s precision and reliability, and effectively adapts the DINO methodology to our specific task. After 200 epochs, a specific stopping criterion was applied, i.e., when the ranking of the top-10 UDs no longer changes, the model is considered to be converged, and training is terminated. The maximum number of epochs was limited to 300 epochs for this study. During the inference mode, only the preprocessing steps on the image resolution and color channels were performed. The embeddings of the lesions were created using the teacher backbone. Using cosine similarity distance, a min max normalized UD score was calculated for each lesion embedding from the median of the lesion embeddings. The top-10 scoring lesions were then proposed as our AI UDs. In \autoref{fig:Figure2}, a complete overview of our proposed approach is shown.

Leveraging the PyTorch implementation provided by the Lightly\cite{lightly} framework with its default model parameterisation, we integrated the previously mentioned modifications of DINO and all model varations such as MoCov2 or different backbones used for the modal comparison. Each individual model, specific to the test patients, was trained on a Tesla®P100 16GB GPU without using any pre-trained weights, generously made available by CSCS. The duration of the training process was on average approximately 30 minutes per patient, predominantly influenced by the number of detected lesions.

\section*{Clinical validation study design}
For the clinical validation study, we instructed participants to identify suspicious-looking lesions on dorsal images of 11 patients. They were asked to prioritize lesions that, in their opinion, needed further examination with a dermascope for melanoma screening, starting with the most suspicious ones. After examining each image, participants rated their confidence on a scale from 1 to 5, with 5 indicating a very confident selection. Unlike other AI studies on the UD sign, our study aimed to include an assessment of the impact of AI as a support tool for screening. For this purpose, participants were asked to repeat the task with the same 11 patient images, this time with the aid of AI predictions. The patient images were displayed alongside their UD scores, with the top-10 AI UDs highlighted in red and the rest in green. However, for the sake of transparency we need to mention that for one patient only the top-9 AI UDs were highlighted in red.

The study aimed to include a diverse range of participants to gauge the impact of AI on decision-making across various pigmented diagnosis expertise levels. Our participant pool consisted of 4 dermatologists with $\leq$ 5 years of experience, 2 dermatologists with $\leq$ 10 years of experience, 3 dermatologists with > 10 years of experience, 2 GP, and 6 non-clinicians, namely non-medical students. This diversity in expertise allowed us to explore how individuals with different backgrounds and levels of experience interacted with AI-driven decision-making processes.

To facilitate the participation in the clinical validation a web tool was developed. The web tool allowed participants to select lesions by drawing bounding boxes around them and included an instructional video at the beginning explaining the task to accomplish. Upon completion of the validation process, it was noticed that some participants had not accurately drawn the bounding boxes around the lesions. As a result, we manually cleaned the data to ensure a correct match between the bounding boxes drawn and the lesions detected by our AI. This correction process was solely undertaken to automate the evaluation, and we took great care to ensure that no corrections influenced the participants’ choices in any way. Subsequently, we created a binary array for each participant, marking the lesions they selected with ones and the others with zeros. Lesions ranked above 20 and those poorly illuminated were excluded from this evaluation. The agreement of our AI algorithms was measured primarily by calculating sensitivity values, defined as follows:

\begin{equation}\label{eq:cvMetrics}
Sensitivity = \frac{TP}{TP+FN}
\end{equation}

Any lesion selected by a participant is considered a UD from their perspective. Thus, True Positives (TP) are the UDs found by the AI algorithm among the top-u ranked lesions. False Negatives (FN), accordingly, are the UDs ranked below the top-u ranked lesions by the AI algorithm or were not detected at all. Similarly, we define the average sensitivity for each participant using the majority voting of experts, namely lesions selected by at least 2 of the 3 dermatologists with >10 years of experience, as the ground truth. For each participant and each patient image, the sensitivity value for the majority selection of experts was calculated and averaged at the end. TP are the lesions that were selected by the participant in question and were part of the respective majority selection of experts. FN are the lesions that were not found by the participant in question and were part of the respective majority selection of experts. The average sensitivity for individual experts with respect to their majority selection was included as well. The absolute values and differences in confidence levels and number of selected lesions with and without the AI-assistance of each participant were additionally evaluated.
\section*{Results}
\subsubsection*{Skin lesion detector}

The YOLOR-P6 model was used to detect skin lesions on four unseen manually labeled patients. The model, with a resolution of 1280 px, achieved an average recall of 95\% and precision of 75\% at an IoU threshold of 50\%. The recall and precision values were calculated for a selected confidence threshold of 20\%. The AP and AR at the same IoU threshold were 95\% and 74\% respectively. When the IoU threshold was increased to 75\%, the AP and AR values slightly decreased to 87\% and 69\% respectively.

Visual inspection revealed that the FN were mainly caused by NMS filtering, which disregarded bounding boxes for lesions close to each other, or freckles that were scored below the 20\% confidence level. Furthermore, False Positives were mainly caused by missing labels for freckles. The precision level for this confidence level is relatively high. Moreover, increasing the IoU threshold shows only a small decrease in the performance metrics. However, there were some patients for whom the model did not perform adequately, namely hairy patients, patients with tattoos, and patients with out of distribution lesions. This was apparent from visual inspection.

\subsubsection*{Ugly duckling detector}
Skin lesion detector sensitivity with respect to ugly ducklings
We further evaluated the sensitivity of our lesion detector by analyzing all selected UDs identified by participants on the 11 patient images. We achieved a recall of 100\% for our lesion detector.

\subsubsection*{Number of ugly ducklings}
\hyperref[fig:Figure3]{\autoref{fig:Figure3}A} shows that dermatologists typically choose on average 4 UDs per patient image without AI assistance. However, there is often significant variability in the number of selected lesions. In contrast, students selected an average of 7 to 8 skin lesions per patient image and even reached the technical limit of 20 lesions on some images. With AI assistance, all participants except GPs and dermatologists with $\leq$ 10 years of experience selected more skin lesions on average for further risk assessment.

\subsubsection*{Top-10 AI sensitivity with respect to participants selection}
In \hyperref[fig:Figure3]{\autoref{fig:Figure3}B}, we show that without any assistance from AI, a top-10 AI sensitivity of 80 to 84\% for dermatologists is achieved, while for the students and GPs lower averages of 63\% and 68\% resulted, respectively. With AI assistance, we achieve high values for dermatologists, ranging from 92 to 98\%, and for students and GPs, we achieved an average sensitivity of 81\% and 82\%, respectively. When provided with AI assistance the interquartile range (IQR) for each group decreased, and for some dermatologists with >5 years of experience even an IQR of zero was achieved.

\subsubsection*{Average participants sensitivity with respect to majority selection of experts}
The agreement among participants with respect to the majority selection of experts was one of the most intriguing questions explored. As shown in \hyperref[fig:Figure3]{\autoref{fig:Figure3}C} dermatologists with $\leq$ 10 years of experience achieved an average sensitivity of 62\% to 66\% without AI assistance, and even decreased slightly with AI assistance for the majority selection of experts. Among the experts themselves, the average sensitivity was 85\% without and 83\% with AI assistance. Students had an average sensitivity of 67\% without and 76\% with AI assistance, while GPs had a relatively low average sensitivity of 57\% without and 51\% with AI assistance. Lastly, we provided the sensitivity values of the top-10 AI UDs. We can see that our AI predictions achieve on average a sensitivity value of 95\% and reach even 100\% on average when presented to the experts. The IQR of each group is quite large in both cases with and without AI assistance. With AI assistance, there is a slight decrease observable for the IQR for dermatologists with $\leq$10 years of experience . The AI predictions have, in both cases, an IQR of zero.

\subsubsection*{Confidence level}
In \hyperref[fig:Figure3]{\autoref{fig:Figure3}D} we show the absolute values found for each participant grouped by expertise. We see a clear difference in the confidence level between students and dermatologists. Interestingly, dermatologists $\leq$ 10 years experience exhibit a strong confidence level compared to the other dermatologists groups. We also noticed that for dermatologists with $\leq$ 5 years of experience the IQR of their confidence decreased to zero after being presented with AI predictions.

In \hyperref[fig:Figure3]{\autoref{fig:Figure3}E}, we additionally show the relative differences in participants’ confidence levels before and after they were exposed to the AI predictions. A clear upward trend in the average is observed for all groups, with the greatest increase being shown by the experts and students. However, for some images, we noticed that the confidence level decreased after participants were shown the AI predictions.

\subsubsection*{Top-u AI sensitivity with respect to participants selection}
\hyperref[fig:Figure3]{\autoref{fig:Figure3}F} displays the average top-u sensitivity of the AI system for different values of u, ranging from 1 to 50, across participant groups. Without the aid of AI, dermatologists achieved average sensitivity values between 65\% and 69\% for the top-5 AI UDs and 89\% for more than 20 lesions. However, on average, students and GPs attained lower top-u AI sensitivity values, but their agreement with AI predictions increased after seeing them. With AI assistance, dermatologists quickly converged to AI sensitivity values above 90\% at u = 8-9. Dermatologists with over 5 years of experience achieved 100\% AI sensitivity at u = 23, whereas those with less than 5 years of experience had slightly lower agreement than their more experienced colleagues. Although the top-u AI sensitivity values for students and GPs with AI assistance were lower than those for dermatologists, there was a clear trend observable that they began to follow the AI predictions more if provided to them.

\subsubsection*{Model comparison of top-u AI sensitivity with respect to majority selection of experts}
\begin{figure}[ht!]
    \centering
    \includegraphics[width=1\linewidth]{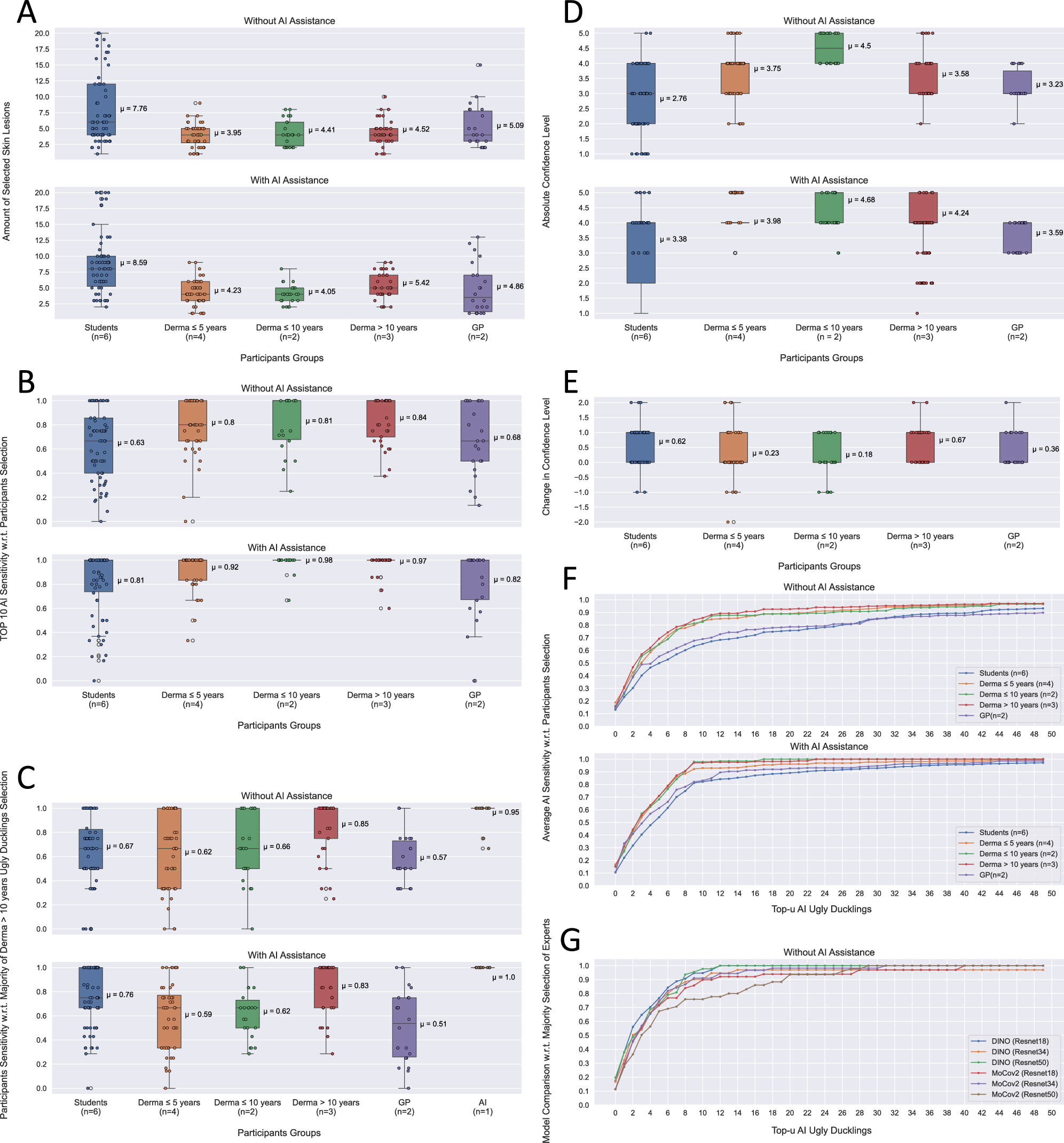}
        \caption{(\textbf{A}) Comparison of the number of UDs selected per patient image and group participant, with and without AI assistance. (\textbf{B}) Comparison of the top-10 AI sensitivity for each patient image with and without AI assistance, with respect to each participant. (\textbf{C}) Average sensitivity values of each group with respect to majority selection of experts, with and without AI assistance. Additionally, top-10 AI sensitivity values averaged over each patient image are also presented. (\textbf{D}) Comparison of the absolute and (\textbf{E}) the relative changes in confidence for participants, with and without AI assistance, across patient images. (\textbf{F}) Average top-u AI sensitivity values for each group, with and without AI assistance. (\textbf{G}) Comparison the average top-u AI sensitivity values for MoCo v2 and DINO models,  with respect to majority selection of experts, for each group without AI assistance.}\label{fig:Figure3}
\end{figure}

In \hyperref[fig:Figure3]{\autoref{fig:Figure3}G}, we compare the sensitivity of our chosen model architecture, DINO, to that of MoCo v2 with respect to the top-u AI sensitivity using majority selection of experts. Both models were tested for the same 3 backbones (ResNet18, ResNet34 and ResNet50). Our analysis shows that DINO outperforms MoCo v2, with DINO achieving over 90\% sensitivity at u=8 for backbones ResNet18 and ResNet50, while MoCo v2 only achieves this level of sensitivity at u=12 for ResNet18 and ResNet34.

\section*{Discussion}
In the present study, a novel approach for total body melanoma screening incorporating self-supervised detection of suspicious lesions based solely on the patient’s context is proposed. The results presented evaluate the impact of this approach on clinicians’ decision-making and confirms its validity through comparison with the assessments of experienced pigmented lesion experts. Total body screening is the most time-consuming and error-prone task, therefore, a reliable support tool would enable dermatologists to better optimize consultation time.

The architecture comprises two main high-level tasks: first, the automatic detection and extraction of lesions from wide field images, and second, the characterization of suspicious lesions through self-supervised clustering. A detailed scheme of the complete pipeline is presented in \autoref{fig:Figure2}. The dataset for this study was collected ad-hoc during routine consultations at the Dermatology Clinic of the USZ and consisted of 90 patients. Due to timing constraints of our clinicians, the proposed AI pipeline was validated only on the dorsal region of 11 randomly chosen patients. To perform the skin lesion detection and extraction, a one-stage object detection model was employed. The model was trained and tested using a semi-automatic process that utilized a combination of a blob detector and manual labeling to accelerate the process on a different randomly chosen data batch consisting of 22 patients. The approach used for lesion detection resulted in high recall rates for all four test subjects, reaching on average 95\% with an IoU threshold of 50\% at 20\% confidence level. This is a critical factor in the detection of potential melanoma cases, ensuring that no lesions are missed. Furthermore, the deep learning model achieved a relatively high Average Precision rate, namely 87\%  $AP_{75}$, while maintaining a high Average Recall rate, namely 69\% $AR_{75}$ , highlighting its effectiveness in detecting skin lesions, even when higher IoU thresholds are used. Worst performance was observed with “atypical” patients, such as those with significant amount of hair. However, in all cases, the suspicious lesions or ugly ducklings (UD) selected by the experienced dermatologist were consistently detected by the object detector. These results reinforce the evidence presented in \cite{PrimaryOBDliterature}, affirming that deep learning techniques are effective in detecting skin lesions.

\begin{figure}[ht!]
    \centering
    \includegraphics[width=1\linewidth]{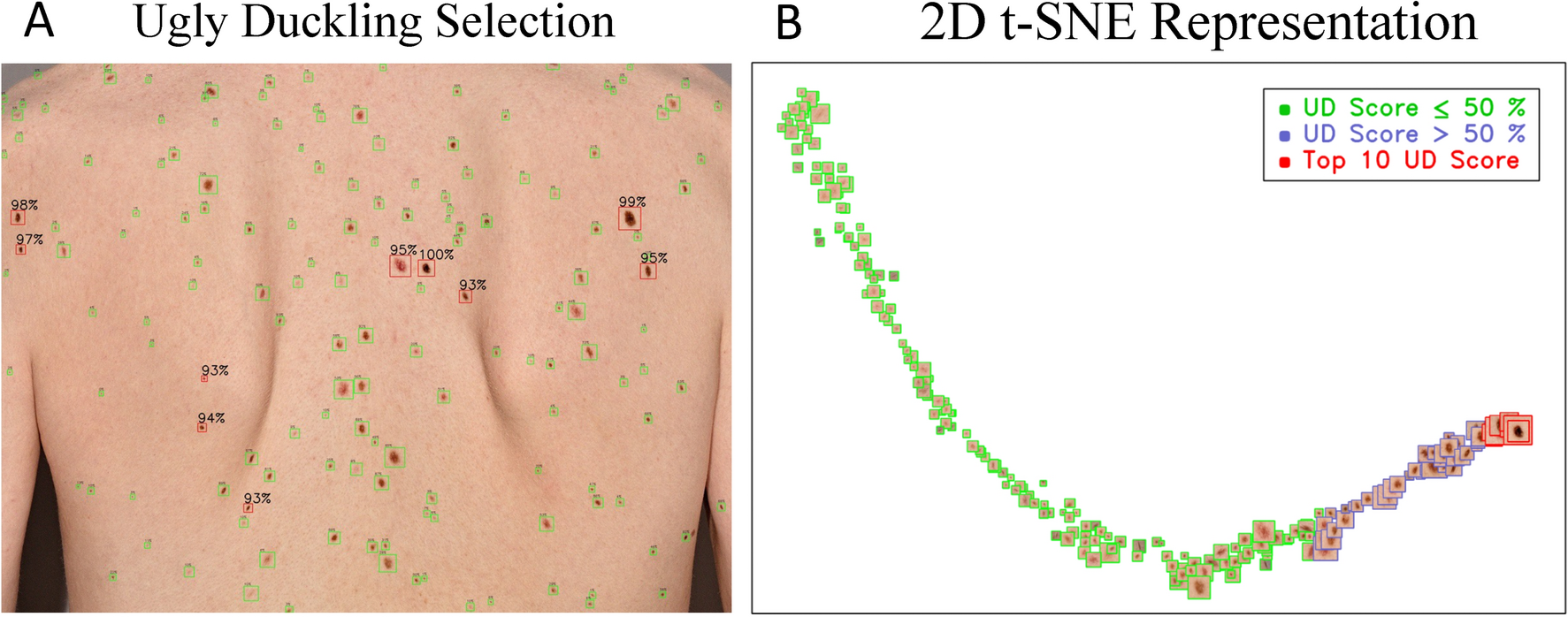}
        \caption{(\textbf{A}) Illustration of the AI’s identification of the top-10 lesions with highest UD scores (red bounding boxes). Above each lesion are the corresponding UD scores is displayed, providing a quantified assessment of the AI’s confidence in each finding. (\textbf{B}) Showing 2D t-SNE visualization of our embeddings in the latent space. With green bounding boxes we see the lesions rating equal or less than 50\% in overall UD score, in blue above 50\% and in red the top-10.}\label{fig:Figure4}
\end{figure}

Regarding the characterization of suspicious lesions or UDs, we propose a self-supervised architecture which allows to evaluate possible outliers considering only the patient’s context. As illustrated in \hyperref[fig:Figure4]{\autoref{fig:Figure4}A}, the algorithm visually presents to the user suspicious looking lesions by drawing a red bounding box around them. The model was able to distinguish UD more effectively compared to common freckles and other average-looking skin lesions. Moreover, visual representation in \hyperref[fig:Figure4]{\autoref{fig:Figure4}B} using t-SNE shows a clear meaningful embedding of UDs versus average-looking lesions.

Despite minor illumination issues that require attention, the overall approach appears to be a promising option. The quantitative comparison of both architectures, DINO and MoCo v2, using 3 different backbones is shown in \hyperref[fig:Figure3]{\autoref{fig:Figure3}G}. We conclude that DINO converged faster towards the mark of 90\% sensitivity for fewer suggested AI ugly ducklings.

Finally, a clinical validation study was conducted to evaluate the performance of our tools and measure its impact in real-world consultation conditions. The study included different groups of interest with varying levels of melanoma screening experience, as introduced in the Clinical Design Study. The study demonstrated that our top-10 AI algorithms performed well for dermatologists, with an average accuracy of 82\%. Furthermore, after reviewing our predictions, dermatologists’ trust in our algorithms increased, reaching an average agreement of 95\%. Therefore, we conclude that this algorithm if deployed in routine consultations with an initial training, nurses or non-experts could reach similar sensitivity values such as experts. Another noteworthy outcome is the number of ugly ducklings selected by experienced dermatologists for each patient. On average they went from 4.52 to 5.42 UDs when being provided by the AI suggestions. This implies that the AI support drew attention to some lesions that were previously overlooked due to their location or the high number of lesions in the image. For two groups, dermatologists $\leq$ 10 years experience and GPs, the number of selected lesions did not increase with AI-assistance. However, they changed their initial selection by following more the algorithm proposals. Inexperienced students, due to their lack of experience and uncertainty with the task, chose more UDs. This also explains their high sensitivity values towards the majority selection of experts and low sensitivity towards AI.

Despite the promising results, we have identified several limitations. Although total body imaging was available for each patient for the sake of proof-of-concept, we restricted ourselves to the dorsal region of 11 patients. Therefore, the next steps should include validating the model in different body regions and on more patients. The challenge of acquiring total body imaging makes our sample size relatively small, which impacts our capacity for evaluating the model’s generalization performance. However, the Dermatology Clinic of the USZ plans to extend the validation campaign to further evaluate the models robustness. Enlarging the number of patients and clinical validation should contribute to increasing the reliability and generalization capacity of our tool. Some patients presented a considerable amount of hair, which impacted both the detection and UD characterization. Despite the limited number of subjects, we should consider a proper way for handling such cases in the future. Additionally, an improved lighting setup will be implemented to prevent shadowed regions in the imaging, thereby avoiding potential detection issues.

\section*{Conclusion}
We propose a novel AI-assisted total body screening tool that achieves expert-level accuracy in identifying suspicious lesions in wide field images, improving upon the results of previous studies ~\cite{doi:10.1126/scitranslmed.abb3652}. The architecture includes a state-of-the-art skin lesion detection system, followed by a self-supervised “ugly duckling” characterization module trained on real-world data acquired from the Dermatology Clinic of the USZ. By eliminating the need for time-consuming manual labeling of UDs and performing predictions on a patient-by-patient basis, we enhance the generalization capacity of our system. These tools ultimately enable the separation of total body screening from routine consultations and even facilitate the involvement of non-expert staff, who can assist dermatologists using reliable tools. The saved screening time can be then reallocated by dermatologists into single lesion assessment and discussions with patients.

\section*{Code availability}
We are unable to release the image dataset used in the study due to the lack of written consent from the patients for data sharing. Code available upon request to PD Dr. Stephanie Tanadini-Lang (Stephanie.Tanadini-Lang@usz.ch) for research purposes.

\section*{Acknowledgements}
This work was supported by the UZH Clinical Research Priority Program “Artificial Intelligence in Radio Oncology” and the Bruno Bloch Stiftung. We would like to thank the Swiss National Supercomputing Centre (CSCS) that provided access to PiZ Daint cluster under project ID “sm71”. Finally, we would like to express our gratitude to all the participants in the clinical evaluation.

\section*{Author contributions statement}
The study was developed within the framework of UZH Clinicial Research Priority Program - Working Package 8 led by JBG and RB. VU, JBG, RB, STL, QL, MB and NA contributed to conception and design of the specifics on the research study presented in this manuscript. VU developed the deep learning pipeline and performed the data analysis with the supervision and collaboration of JBG. VU and JBG wrote the first draft of the manuscript. All authors contributed to manuscript revision, read, and approved the submitted version. Funding acquisition and project administration was carried out by RB, STL and NA.

\section*{Competing interests}
The authors declare no competing interests.

\bibliography{sample}



\end{document}